\documentclass[letterpaper,12pt, abstracton]{scrartcl}

\usepackage{marvosym}
\usepackage[T1]{fontenc}
\usepackage{color,hyperref}
\usepackage{enumitem}
\usepackage{amssymb}
\usepackage{amsmath}
\usepackage{pifont}
\usepackage{makecell}
\usepackage{subcaption}
\usepackage{booktabs}
\usepackage{Sweave}
\date{}

\begin{document}
	
\title{IgNet. \\ A Super-precise\\ Convolutional Neural Network.}
\author{Igor Mackarov \\ \href{mailto:Mackarov@gmx.net}{{\small Mackarov@gmx.net}}}

\maketitle
\begin{abstract}

Convolutional neural networks (CNN) are known to be an effective means to detect and analyze images. Their power is essentially based on the ability to extract out images\textquotesingle~common features. There exist, however, images involving unique, irregular  features or details. Such is a collection of unusual children drawings reflecting the kids\textquotesingle~imagination and individuality. These drawings were analyzed by means of a CNN constructed by means of Keras--TensorFlow. The same problem --- on a significantly higher level --- was solved with newly developed family of networks called IgNet that is described in this paper. It proved able to learn by 100\% all the categorical characteristics of the drawings. In the case of a regression task (learning the young artists\textquotesingle~ages) IgNet performed with an error of no more than 0.4\%. The principles are discussed of IgNet design that made it possible to reach such substantial results with rather simple network topology.


\end{abstract}

\section{Introduction}
\label{Intro}

Convolutional neural networks (CNN) are known to be an effective means to detect and analyze images. Their power is essentially based on the ability to extract out images\textquotesingle~regular features. Changing learning depth, or the number of convolutional layers, lets distinguish the features with various levels of abstraction. Stacking \textit{convolutional} layers together with \textit{pooling} ones can bring about \textit{translation invariance} (when an exact position of a feature is irrelevant) \cite{DeepLearning} thus reaching state-of-the-art accuracy.

There exist, however, tasks involving unique, irregular images lacking common features or details. Such is a psychological collection of children\textquotesingle s drawings available on the site of University of Stirling~ \href{http://pics.psych.stir.ac.uk}{http://pics.psych.stir.ac.uk}. The collection\textquotesingle s uniqueness is in detail discussed in section~\ref{sec:Peculiarities}.

The work presents results of two solutions to the regression---classification problem. 

The first one, fairly precise, with very thorough choice of the CNN topology, was obtained using Keras--TensorFlow.

 The second one was achieved by a new CNN system written from the scratch on C++ and called \textit{IgNet}. It proved possible to reach very high accuracy (in particular, 100\% in all the classification configurations tried) even with simplest network structures. Section \ref{sec:IGNET} gives the details of that, specifically pointing out the IgNet details that were crucial in reaching such performance.

\section{Description of the drawings database}

\subsection{The labels}
\label{sec:labeks}

\begin{figure}[ht]
	\centering	
	\includegraphics[width=0.5\textwidth]{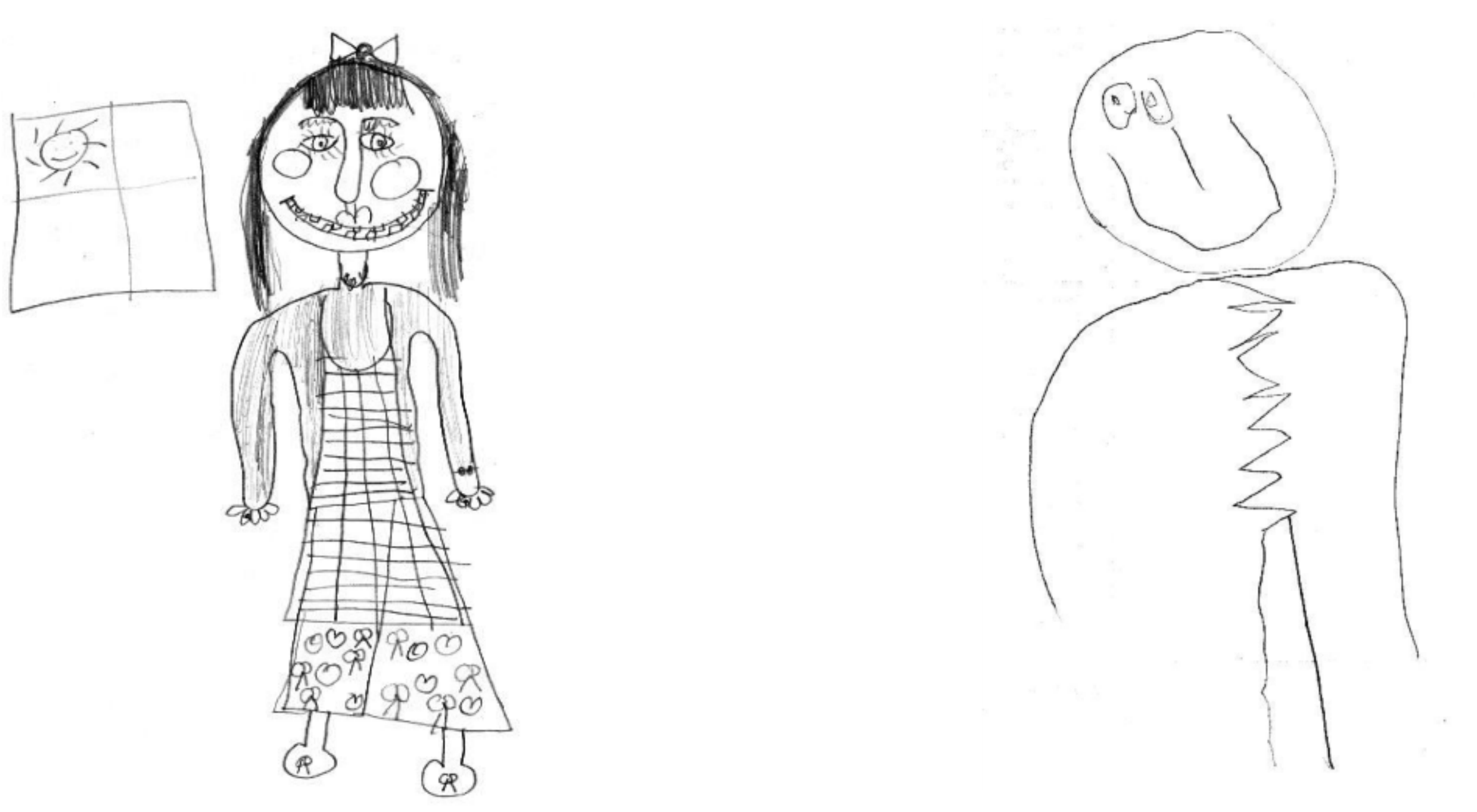}\\
	\caption{Two samples from the kids drawings collection}
	\label{fig:twokids}
\end{figure} 

The set of kids drawings is collected as a result of a psychological research in the form of jpeg files whose names contain a number of  encoded categorical and numerical data about  a drawing and its author. For example, the left-side picture on Fig.~\ref{fig:twokids} is called \\ \textbf{\textit{p3-67w-6.7f.jpeg}}, which means:

\begin{itemize}
	\item age category -- p3,
	\item \underline{age -- 6 years, 7 months},
	\item author -- a girl,
	\item who is drawn -- a female.	
\end{itemize}	

The right-side picture is \textbf{\textit{nu-21s-4.5m.jpeg}}, thus

\begin{itemize}
	\item age category -- nursery,
	\item \underline{age -- 4 years, 5 months},
	\item author -- a boy, \textit{self-portrait},
	\item who is drawn -- a male.	
\end{itemize}	

We therefore have three groups of categorical data and the fourth group (underlined) that can be expressed as \textit{the age in months}. It is possible to treat this group as either categorical or numerical (a learned age of, for example, 53.4 months certainly makes sense).

\section{Peculiarities of the drawings}
\label{sec:Peculiarities}

\subsection{Uniqueness}

The value of regular features for successful analysis of an image was emphasized above. For example, human faces, in all their variety, have easily recognizable details such as eyes, noses, etc. Bodies are recognized by heads, limbs... Yet, Fig.~\ref{fig:twokids} shows how different the drawings of children of different ages are, and how few regular and recognizable components they have in common. Such are nearly all the drawings in the database. They are unique and reflect the kids\textquotesingle~imagination and individuality.

\subsection{Multifactoriality}

To get the most valuable information, do we need to fit the model using each of the available labels   separately, or use them together simultaneously? Subsections \ref{sec:encoding}, \ref{sec:preprocessing} explain the decisions taken.

\subsection{Small size of the database}

The data collection has only 338 drawings. As a rule, any Data Science model can be successfully learned on a database consisting of at least several thousands of samples. How the database is enlarged, is described in subsections \ref{sec:augmentation} and \ref{sec:preprocessing}.\\

\section{The Keras Solution}
\label{sec:KerasSolutionn}

\begin{figure}[h]
	\centering	
	\includegraphics[width=1\textwidth]{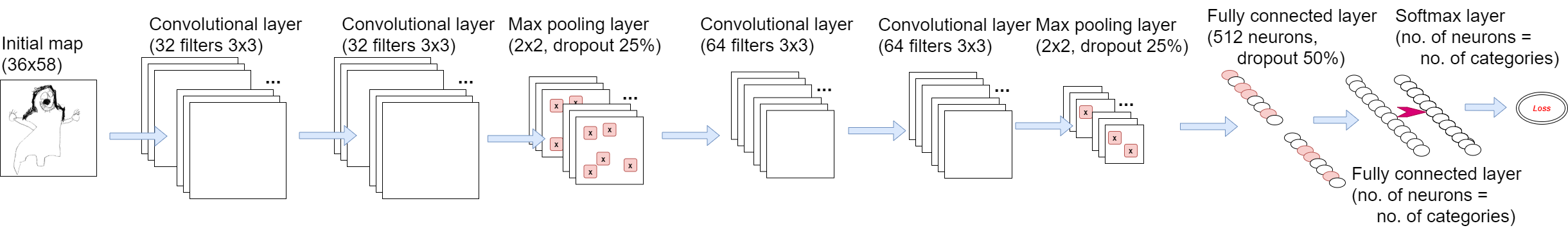}\\
	\caption{The Convolutional Neural Network constructed within the Keras framework to solve the problem involved. Classification node. }
	\label{DeeperNet}
\end{figure}

The key to a successful problem solving was construction of a right network. The constructed CNN (Fig.~\ref{DeeperNet}) is significantly deeper compared to largely known \textit{LeNet} \cite{Heaton1, LeNet1, LeNet2, LeNet3} that has two pairs of convolutional and pooling layers and, unfortunately, failed to learn in our situation. 

The constructed CNN includes two blocks, each in its turn consisting of a pair of convolutional layers followed by \textit{RELU} activation, and one \textit{max-pool} layer. Two fully connected layers with \textit{sigmoid} activation follow them. In the case of classification their output is modified by a \textit{softmax} layer. This makes it possible to interpret the network output as probabilities of belonging the input image to each of the categories. Finally, loss functions are \textit{cross-entropy} and \textit{MSE} for classification and regression, respectively.

Another characteristic feature of the network is use of \textit{dropout} in \textit{max-pool} and one of the fully connected layers to fight overfitting.

\subsection{Images contraction}
\label{sec:contraction}
Following best practices, for faster calculations we reduced the images from their initial sizes (about 600$\times$800 pixels) to 36$\times$58 pixels using the Python \textit{cv2} library.
  
\subsection{Database augmentation}
\label{sec:augmentation}

One of the ways to deal with small datasets is `reproduction' of visual data by stochastic rotations--shifts--flips of images horizontally and vertically. With \textit{Keras}  \textit{ImageGenerator} utility the drawings set was enlarged up to 17500 items for learning and testing. 3500 more samples were produced for validation.

\subsection{One-hot encoding}
\label{sec:encoding}

The problem of multifactorial classification was solved by \textit{one-hot encoding} approach with use of \textit{Keras} API (the \textit{to\_categorical} method). 
The opportunity was provided to learn either on each of the four groups separately, or on any possible combination of those. In the case of all four groups involvement, the whole class consisted of 186 categories.

Note that all the variants gave comparable accuracies. This is surprising, for preliminary it was natural to expect that an image is most precisely identified by all the factors. In fact, it turned out that each of the four factors or their possible combinations are equally significant and able to classify a drawing (\textit{vide} Table~\ref{tbl:res}).

\subsection{Details of learning}

\subsubsection{Ten-fold cross-validation}
\label{sec:Ten-fold}

As was noted in subsection \ref{sec:augmentation}, the whole of the augmented dataset was divided into \textit{learn-test} and \textit{validation} parts. The former, in its turn, was split in 10:1 ratio to provide data for learning and testing, respectively. This was repeated ten times to produce ten partitions like this, so that each sample could be used for testing once and only once per a whole dataset processing. After a cycle of learning and testing with all ten folds a \textit{validation} was performed on the appropriate part of the common dataset, whose elements never participated in the learning--testing procedure.

A version of \textit{stratified cross validation} was used: in each fold, proportions of samples with any of the categories coincided with those in the whole of the \textit{learning-testing} dataset. This was provided by Python\textquotesingle s \textit{sklearn}  library (the \textit{KFold} method).

\subsubsection{Early stopping}

\textit{Overfitting} is a common issue of various Machine Learning models: starting from certain level of learning-testing accuracy the \textit{generalization}, i.~e., accuracy of classification or regression with data outside a learning--testing dataset, can start to decay. In this solution we abandoned a next cycle of learning on a current fold as soon as validation accuracy decay was observed a few times in a row. This number of times, the \textit{tolerance}, was chosen 7, with the number of consecutive processing of a folder 10.

\subsection{The results}

Thus, two tasks were solved on 17500--sample learning--testing dataset and 3500--sample validation one. 
\paragraph{Regression.}  Age as a numeric value was predicted with MAPE (mean absolute percentage error) \textit{\textbf{12.95\%}}.

\begin{table}[th]
	\centering	
	\caption{Keras--TensorFlow solution: classification accuracy (\textit{percentage of right guesses about the categories}) for various combinations of factors.}
	\label{tbl:res}
	\begin{tabular}{@{} l *4c @{}}
		\toprule
		\multicolumn{1}{l}{Factors included}    & Accuracy    \\ 
		\midrule
		`age', `who drew', `who is drawn', `age category' & 85.58\% \\ 
		`who drew', `who is drawn', `age category'  & 77.80\%  \\
		`who is drawn', `age category'  & 83.72\%  \\
		`age category'  & 88.54\%  \\
		\bottomrule
	\end{tabular}
\end{table}

\paragraph{Classification.} As was pointed in subsection \ref{sec:encoding}, learning  on one, two, three, or all four categories brought about comparable accuracies, which is illustrated by Table \ref{tbl:res}.\\\\

\section{The IgNet Solution}
\label{sec:IGNET}
However solid the above shown results are, still better solution was obtained with a new family of network IgNet. 

It is written on C++ from the scratch and makes use of a number of modern proven techniques that boost performance in neural networks. The high rate of calculations is reached by dint of multithreading on both central and graphical processing units (CPU, GPU - subsection~ \ref{sec:GPU}). Another resource greatly improving the performance is a possibility to fine-tune the feed-forward --- back-propagation convergence.

Describe IgNet and how it works step by step.

\begin{figure}[!t]
	\centering
	\begin{subfigure}[]{0.25\textwidth}
		\includegraphics[width=1\textwidth]{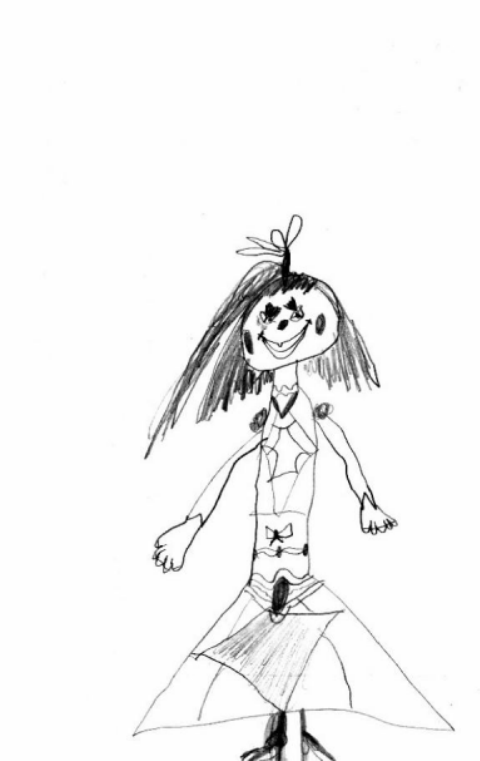}
		\caption{initial picture}
	\end{subfigure}
	\begin{subfigure}[]{0.25\textwidth}
		\includegraphics[width=1\textwidth]{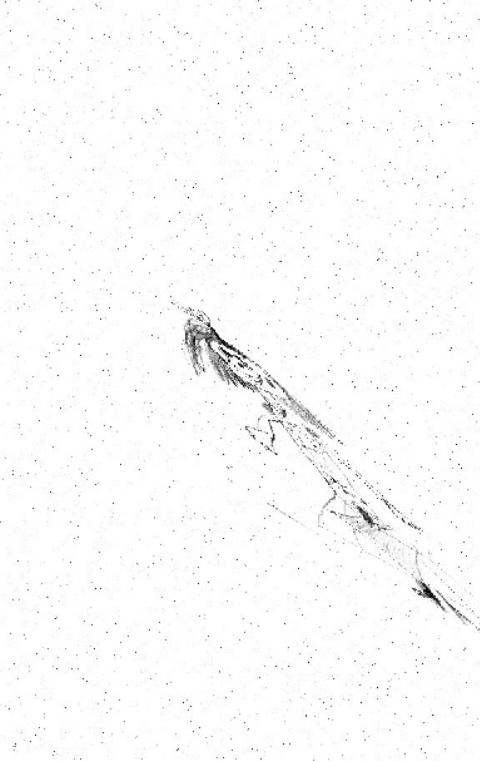}
		\caption{transformed image}
	\end{subfigure}
	
	\begin{subfigure}[]{0.25\textwidth}
		\includegraphics[width=1\textwidth]{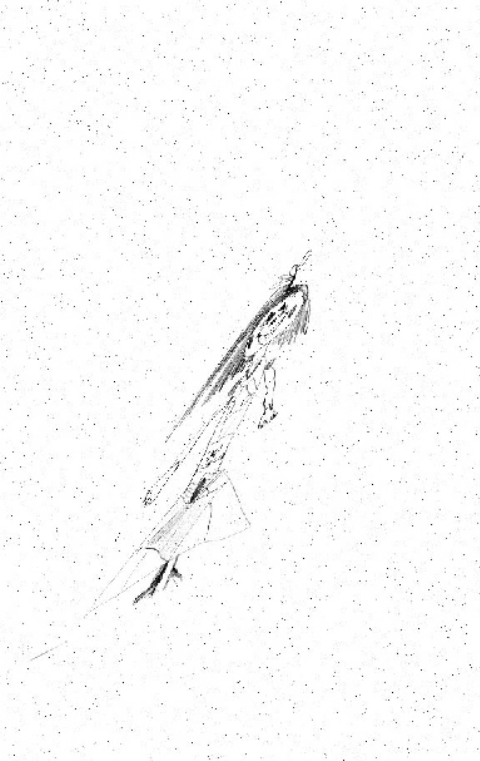}
		\caption{transformed image}
	\end{subfigure}
	\begin{subfigure}[]{0.25\textwidth}
		\includegraphics[width=1\textwidth]{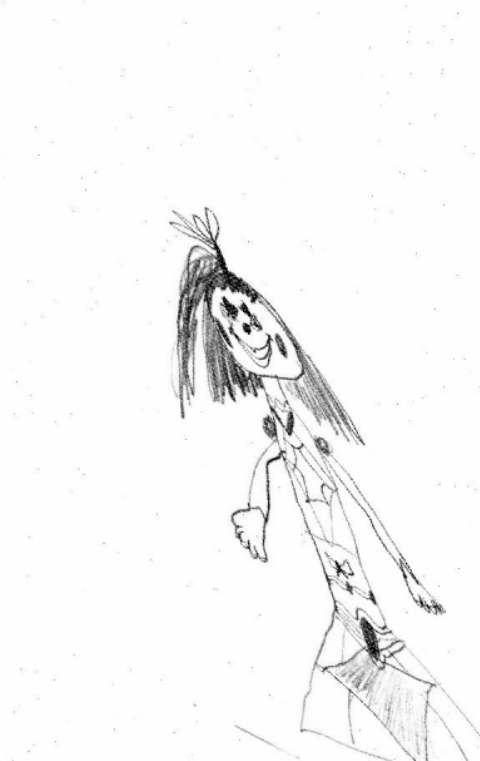}
		\caption{transformed image}
	\end{subfigure}
	\caption{Augmentation of an initial picture by means of random rotation, stretch in both directions, and noising.}
	\label{fig:augmentation}
\end{figure}

\subsection{Database preprocessing}
\label{sec:preprocessing}

\paragraph*{Augmentation.} Instead of using standard Python--Keras means, we augmented the given dataset on the fly by the functionality that is a part of IgNet. In addition to linear transformations of images, it provides \textit{noising}, i.~e., adding random positive or negative values to randomly chosen pixels (\textit{vide} Fig.~\ref{fig:augmentation}). This was controlled by setting relative level of noise and portion of pixels to be noised. As one of the augmentation methods, noising proved the most effective way to make an extended dataset learn fast. Unlike the Keras solution, where the dataset was augmented by nearly thirty times, only ten times augmentation turned out to be enough for a successful learning.

\paragraph{One-hot encoding} was also performed on the fly within the framework of the classification process. Overall, \textit{`age in months'} as well as each one and every possible combination of the	\textit{`age', `who drew', `who is drawn'}, and  \textit{`age category'} labels (cf. subsection \ref{sec:labeks}) were considered.

\subsection{Details of learning--testing--validation}

 They were pretty close to the Keras case. Here the network also learned on each fold ten times in a row with tolerance  7. Unlike the Keras solution, the number of folds was 5.

\subsubsection{Gradient descent} 
\label{sec:descent}

As in many cases elsewhere and in the Keras case, \textit{mini-batch gradient descent} proved most effective (compared to stochastic and full gradient descents). The batch size was chosen \\to be 32. As to the descent technique, together with the simple one, Momentum and Nesterov accelerated gradient (NAG) were used \cite{Dangeti}. With respect to this particular dataset, those two did not give any remarkable acceleration compared to the traditional form.

\subsection{Higher rate of computations due to CPU multithreading and GPU}
\label{sec:GPU}

The feed-forward stage of the CNN learning process is known to consist of a number of independent `elementary convolutions' so it could be quit natural to organize computations in the form of \textit{parallel threads} or \textit{work items} in the case of GPU.

\paragraph{The GPU mode}was implemented using \textit{OpenCL} (a C wrapper for GPU \cite{GPU}).
\paragraph{The CPU multithreaded mode}was designed in the form of a pull of threads starting once at the beginning of the calculations and periodically fulfilling their assignments. The most challenging point here was to make the threads simultaneously resume at a specified moment (beginning of the convolution stage) and get asleep as soon as each one is done with its assignment (a sequence of races starting at a specified moment and ending when all the participants finish). This was implemented by means of C++ STL \textit{lockable mutexes} and \textit{condition variables} \cite{CPU}.

\subsection{Control of divergence by initial weights distribution range}

 Need in accurate choice of gradient descent technique was discussed in Subsection \ref{sec:descent}. Generally, it aims at reaching stable and fast convergence of learning process. Not only does convergence depend on the gradient descent technique, but also on the qualities of \textit{the loss function surface}, its steepness, in particular. The latter depends on values of partial weights derivatives of the loss function. When, except weights, biases learn too, convergence also depends on its biases partial derivatives.

\subsubsection{Formal analysis of a CNN back propagation convergence abilities}

According to the definition of a convolutional neural network \cite{DeepLearning}, a partial derivative of loss function with respect to the \textit{ij}-th weight of \textit{the last layer} (i.e., the \textit{ij}-th element of any filter of the last layer) can be expressed as follows:

\begin{equation}
{\left( {\frac{{\partial {Loss}}}{{\partial {w_{ij}}}}} \right)_{last}} = \sum\limits_{\beta\left(i,j\right)} {\left( {In_{\alpha\left(\beta\right)}} \right)_{last} \left( {N'_\beta} \right)_{last} \left( {\frac{{\partial Loss}}{{\partial N_{\beta\,\,last}}}} \right)}.
\label{eq:der_w_last}
\end{equation}
\noindent
Here:
\begin{itemize}
	\item $In$ is an element of the last layer input map,
	\item $\alpha$ designates all the input map elements being multiplied by $w_{ij}$ in convolution for the position of the weights filter corresponding to neuron $N_\beta$ \footnote{Explain presence of $\beta$ in Eq.~\eqref{eq:der_w_last}: IgNet\textquotesingle s last layer is able to have more than one neuron, even a matrix of neurons.},
	\item ${N'}$ is the derivative of an activation function of a neuron (of the last layer output map element), 
		\item $\beta$ labels all neurons of the last layer that are being activated by convolutions with participation of $w_{ij}$.
\end{itemize}

For any other (the $n$-th)  layer we have 
\[
{\left( {\frac{{\partial {Loss}}}{{\partial {w_{ij}}}}} \right)_n} = \sum\limits_{\alpha,\beta,\gamma,\delta} {\left( {In_{\alpha\left(\beta\right)}} \right)_n \left( {N'_{\beta(i,j)}} \right)_n \left( w_{\gamma\left(\beta\right)} \right)_{n + 1} \left( {N'_{\delta\left(\beta\right)}} \right)_{n + 1}},\,\,\,\,\,\, n=1\div last-1, 
\tag{1a}
\label{eq:der_w_nth}
\]
where:
\begin{itemize}
	\item $In$ is an element of the \textit{n}th layer input map,
	\item $\alpha$ terms all the input map elements being multiplied by $w_{ij}$ in convolution for the position of the weights filter corresponding to neuron $N_\beta$,
	\item ${N'}$ is the derivative of an activation function of a neuron ( element of output map of the $n$th or the ($n+1$)-th  layer), 
	\item $\beta$ points to all the \textit{n}th layer neurons that are being activated by convolutions with participation of $w_{ij}$,
	\item $\gamma$ indexes all the ($n+1$)-th  layer weights being multiplied by neurons $N_\beta$ as elements of the ($n+1$)-th layer input map,
	\item $\delta$ accounts	for all the neurons that are being activated with participation of these weights.
\end{itemize}

Regard two last sum terms in Eq.~\eqref{eq:der_w_last} and three last terms in Eq.~\eqref{eq:der_w_nth}. Evidently, they present \textit{a full derivative} of the loss function over the $ij$-th bias (the one participating in  the convolution activating the $ij$-th neuron of the last or $n$th layer):

\begin{equation}
{\left( {\frac{{\partial {Loss}}}{{\partial {b_{\beta}}}}} \right)_{last}} =  {\left( {N'_\beta} \right)_{last} \left( {\frac{{\partial Loss}}{{\partial N_{\beta\,\,last}}}} \right)} ,
\label{eq:der_b_last}
\end{equation}

\[
{\left( {\frac{{\partial {Loss}}}{{\partial {b_{\beta}}}}} \right)_n} = \sum\limits_{\beta ,\gamma, \delta } {\left( {N'_\beta} \right)_n (w_{\gamma(\beta)})_{n + 1} \left( {N'_{\delta(\beta)}} \right)_{n + 1} } ,
\tag{2a} 
\label{eq:der_b_nth}
\]

\noindent
the notation is the same as in Eqs.~\eqref{eq:der_w_last}--\eqref{eq:der_w_nth}.

The above expressions turned out to be useful for control of the loss function weights derivatives and its surface steepness. Practical adjustment of a loss function will be exemplified in the next subsection.

\paragraph{Note:} followed \textit{a softmax layer}  the last one, Eqs.~\eqref{eq:der_w_last} and \eqref{eq:der_b_last} would be some more complicated; we omitted these forms here, but principal dependencies and suggestions of subsection \ref{seq:esteem} remain the same for this case.

\subsubsection{Practical improvement of convergence}
\label{seq:esteem}

Describe \textbf{a scheme} of boosting the rate of learning on the basis of expressions \eqref{eq:der_w_last}~and~\eqref{eq:der_w_nth} which we used in this research. 

First of all, the samples were normalized to be between 0 and 1. Further on, initial \textit{uniform}  distributions of weights and biases for all the layers of a CNN were symmetric. Weights were normalized on product of the filter vertical and horizontal spacial extents:

\begin{equation}
w_{ij}^n \in \left[ { - \frac{{W_{}^n}}{{v^nh^n}},\,\,\,\,\,\frac{{W_{}^n}}{{v^nh^n}}} \right],\,\,\,
b_{ij}^n \in \left[ - B_{}^n,\,\,\,\,\,B_{}^n\right],\,\,\,\,\,\,\,\, n=1\div last.
\label{eq:distr_b}
\end{equation}

\noindent
For a $kl$-th element of the $n$th layer input map, which is the result of $n-1$ convolutions, we then have:
\begin{equation}
\left| {{{(I{n_{kl}})}_n}} \right| \le \prod\limits_{\nu  = 1}^{n-1} {({B^\nu} + {W^\nu})},\,\,\,\, n=2\div last
\label{eq:In_nth}
\end{equation}

\noindent
To make the basic relations easier to use, we applied \textit{identity} activation functions in all layers to have the  derivatives of neurons in \eqref{eq:der_w_last}--\eqref{eq:der_w_nth} equal 1. In subsection \ref{sec:why} we will discuss more advantages of getting rid of non-linear activations. For a CNN with layers $1\div last$ from Eqs.~\eqref{eq:der_w_last}--\eqref{eq:der_w_nth} and \eqref{eq:In_nth} it follows

\begin{equation}
{\left|\left( {\frac{{\partial {Loss}}}{{\partial {w_{ij}}}}} \right)_{last}\right|} \le \mathop {max\left(\left| {\frac{{\partial Loss}}{{\partial N_{\beta\,\,last}}}} \right|\right)\,}\limits_\gamma \prod\limits_{\nu  = 1}^{last-1} {({B^\nu} + {W^\nu})},
\label{eq:non-eq-last}
\end{equation}

\[\left| {{{\left( {\frac{{\partial Loss}}{{\partial {w_{ij}}}}} \right)}_n}} \right| \le \left\{ {\begin{array}{*{20}{c}}

	{\mathop {max({W^{n + 1}}){\mkern 1mu} }\limits_\gamma  \prod\limits_{\nu  = 1}^n {({B^\nu } + {W^\nu })} }	{{\quad:\quad}}&{n = 2\, \div \,last - 1}\\
	{\mathop {max({W^2}){\mkern 1mu} }\limits_\gamma  }	{{\quad:\quad}}&{n = 1.}
	\end{array}} \right.
\tag{5a} 
\label{eq:non-eq-nth}
\]

\noindent
Since we \textit{randomly} set initial weights using uniform distributions, the weights derivatives calculated \textit{after one feed forward and one backpropagation} have equal chances to take any value between zero and right-hand sides of expressions \eqref{eq:non-eq-last} and \eqref{eq:non-eq-nth} that are bounded\footnote{For expression \eqref{eq:non-eq-last} this is true because derivatives of traditional loss functions (like MSE and MAE) on output neurons are bounded. This is also true in case of \textit{softmax -- cross-entropy} combination; one can easily get convinced of this, having noticed that the softmax derivative with respect to output neurons \\${\raise0.7ex\hbox{${\partial {\bf{s}}}$} \!\mathord{\left/
			{\vphantom {{\partial {\bf{s}}} {\partial {N_i}}}}\right.\kern-\nulldelimiterspace}
		\!\lower0.7ex\hbox{${\partial {N_i}}$}} = {s_i}(1 - {s_i})$, whereas~~ ${\raise0.7ex\hbox{${\partial CrossEntropy}$} \!\mathord{\left/
			{\vphantom {{\partial CrossEntropy} {\partial {s_i}}}}\right.\kern-\nulldelimiterspace}
		\!\lower0.7ex\hbox{${\partial {s_i}}$}} $  is ${\raise0.7ex\hbox{$1$} \!\mathord{\left/
			{\vphantom {1 {{s_i}}}}\right.\kern-\nulldelimiterspace}
		\!\lower0.7ex\hbox{${{s_i}}$}}$ when $target_i=1$ and 0 when $target_i=0$.}. Evidently, after repetition of these two processes from the beginning a due number of times with varying amplitudes $B^\nu > 0$, $W^\nu > 0$ (if necessary) we can come to \textit{acceptable} values of the derivatives. The number of pairs $B^\nu$ --- $W^\nu$ equals the number of the layers, so  the loss function partial derivatives with respect to the weights of each layer can be adjusted independently. 

As the practical experience suggests, orders of the weights and derivatives at the beginning of learning do not significantly change further on it. Nor do they greatly depend on the variety of input samples. So the derivatives estimation at the start, as a rule, remains correct for any later moment. Regarding the problem solved,  \textit{acceptable} derivatives were of order of 0.01 --- 0.1. Smaller values most often resulted in very slow learning. Higher values could invoke a burst of instability (gradient explosion).

\begin{figure}[!t]
	\centering
	\begin{subfigure}[b]{0.95\textwidth}
		\includegraphics[width=1\linewidth]{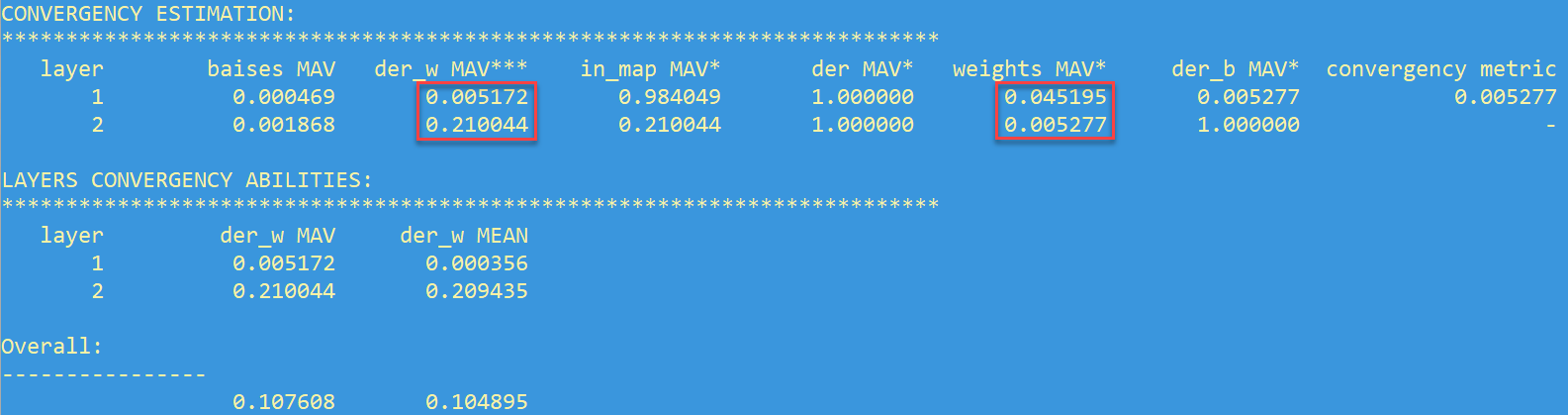}
		\caption{Too small weights for the second layer were specified, which resulted in small weights derivatives for the first layer and very poor convergence.}
		\label{fig:Ng1} 
	\end{subfigure}
	
	\begin{subfigure}[b]{0.95\textwidth}
		\includegraphics[width=1\linewidth]{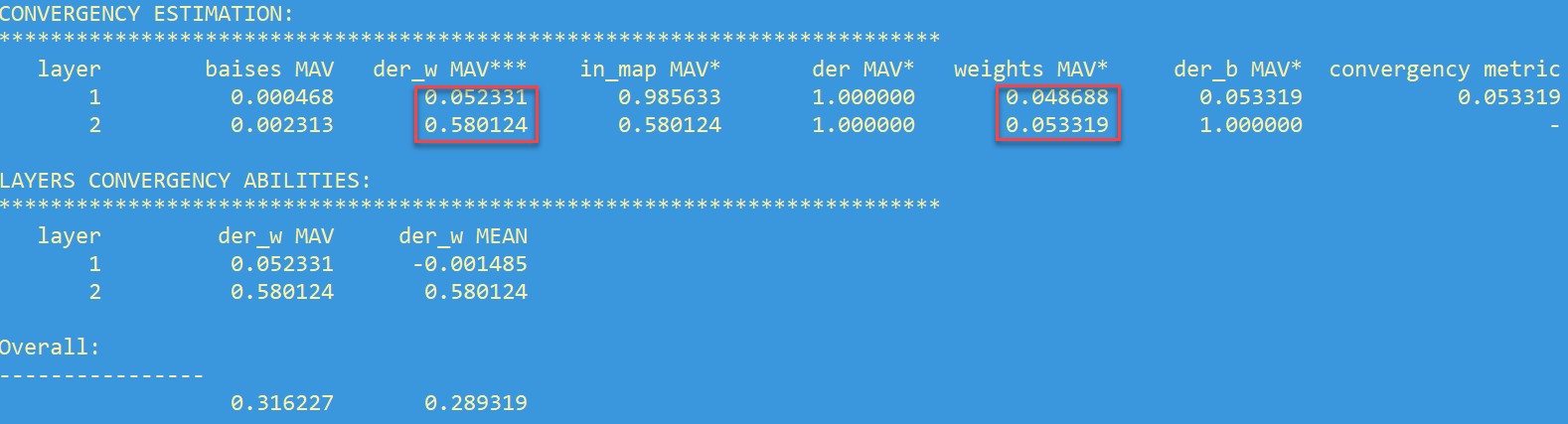}
		\caption{Bigger weights for the second layer specified. This resulted in rise of the first layer weights derivatives MAV by an order. Fast learning followed that with testing and validation MAPE about 0.4\%. }
		\label{fig:Ng2}
	\end{subfigure}
	
	\caption{Console output of a two-layer IgNet (regression) at the beginning of the learning. It assesses expected rate of convergence. Marked with an asterisk are MAV of a convolutional layer input map, of a neuron activation derivative, of a layer weights, of a full neuron derivative with respect to a bias. Marked with three asterisks is MAV of the loss function weights derivatives  that depend on the specified values as per Eqs.~\eqref{eq:der_w_last}~and~\eqref{eq:der_w_nth}. }
	\label{fig:console}
\end{figure}

\begin{figure}[!t]
	\centering	
	\includegraphics[width=0.7\textwidth]{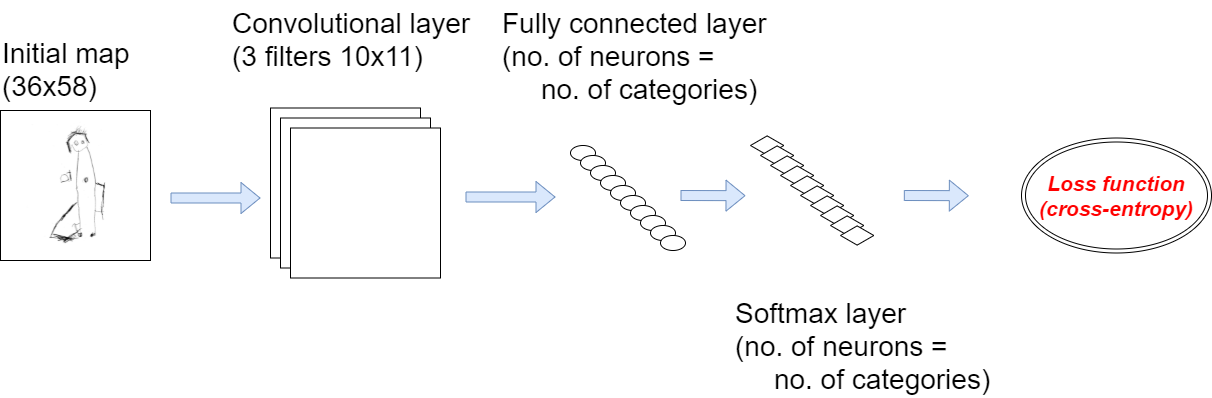}\\
	\caption{One of the simplest CNN constructed within the framework of IgNet to solve the problem involved. Classification mode.}\label{fig:ShallowNet}
\end{figure}

\begin{figure}[!h]
	\centering
	\includegraphics[width=1\textwidth]{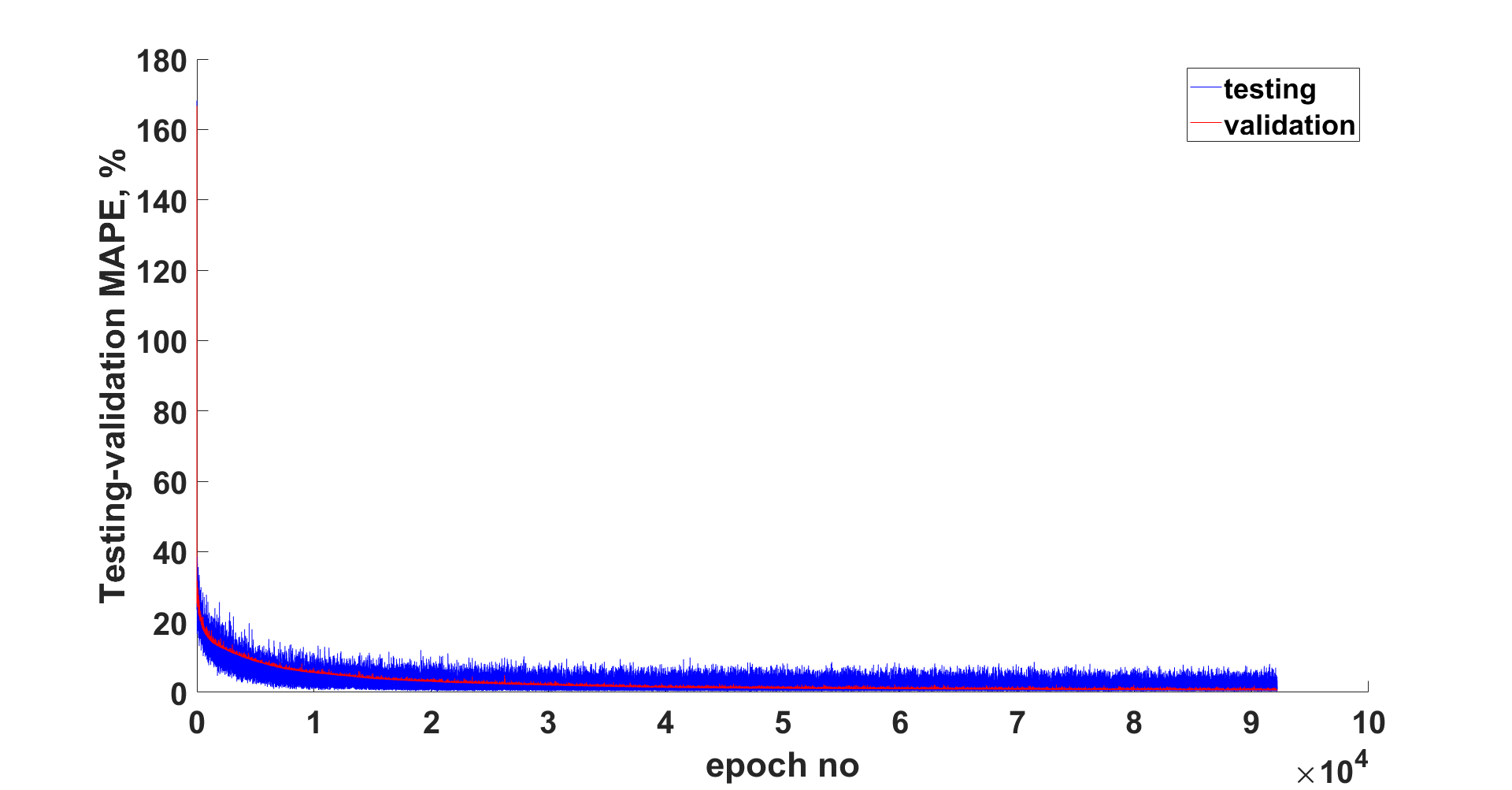}
	\caption{The performance of the Fig.~\ref{fig:ShallowNet} network. Learning profiles for the regression on the `age' target.}
	\label{fig:regression}
\end{figure}

Figure \ref{fig:console} illustrates how the scheme described is applied to the simplest network shown in Fig.~\ref{fig:ShallowNet}. At the beginning of the learning, upon the first backpropagation (or possibly at any other moment) IgNnet puts out \textit{mean absolute values} (MAV) of the variables present in Eqs.~\eqref{eq:der_w_last}~and~\eqref{eq:der_w_nth} to console. Having noticed that MAV of Loss derivatives with respect to the first layer weights is too small, we still go on with the learning, which, even upon a long time, gives rather significant testing and validation errors. Then we restart the calculations with the second layer weights distribution amplitude increased by 10 times. In Fig.~\ref{fig:Ng2} we see that the Loss derivatives on the first layer weights have increased approximately by an order --- just in compliance with Eq.~\eqref{eq:der_w_nth}. This is a good news, because the further learning will give really nice results (\textit{vide} Fig.~\eqref{fig:regression}).

\noindent
\paragraph{Note:}Eqs.~\eqref{eq:der_b_last}--\eqref{eq:der_b_nth} can generally be used for estimation of the loss derivatives with respect to the biases in case of biases teaching. For the task involved it was decided, however, not to teach biases at all. Control of weights distributions only, gave simpler dependencies and eventually better performance.

\subsection{The results and discussion}

As per subsection \ref{seq:esteem}, with accurate choice of weights range a network with just one convolutional and one fully connected layers proved able to very precise learning. Such precision is far beyond the abilities of much more sophisticated network shown in Fig.~\ref{DeeperNet} constructed with Keras--TensorFlow.

\subsubsection{Successful networks structures}
\begin{figure}[!h]
	\centering	
	\includegraphics[width=1\textwidth]{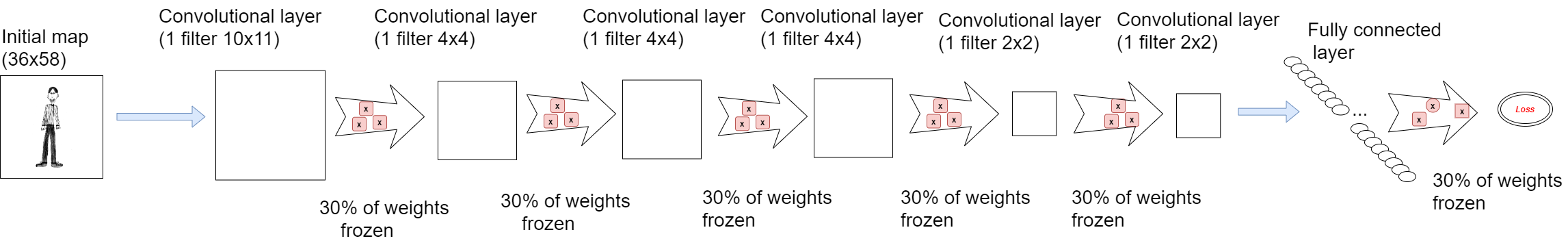}\\
	\caption{The deepest CNN constructed within the framework of IgNet to solve the problem involved. Regression mode.}
	\label{fig:DeepestNet}
\end{figure}

At the same time, for the sake of better understanding of IgNet\textquotesingle s abilities, other networks were tried.
Namely, networks with 3, 4, 5, 6, and 7 layers. The latter, the most complex, is illustrated by Fig.~	\ref{fig:DeepestNet}.

\begin{figure}[!t]
	\centering
	\includegraphics[width=1\textwidth]{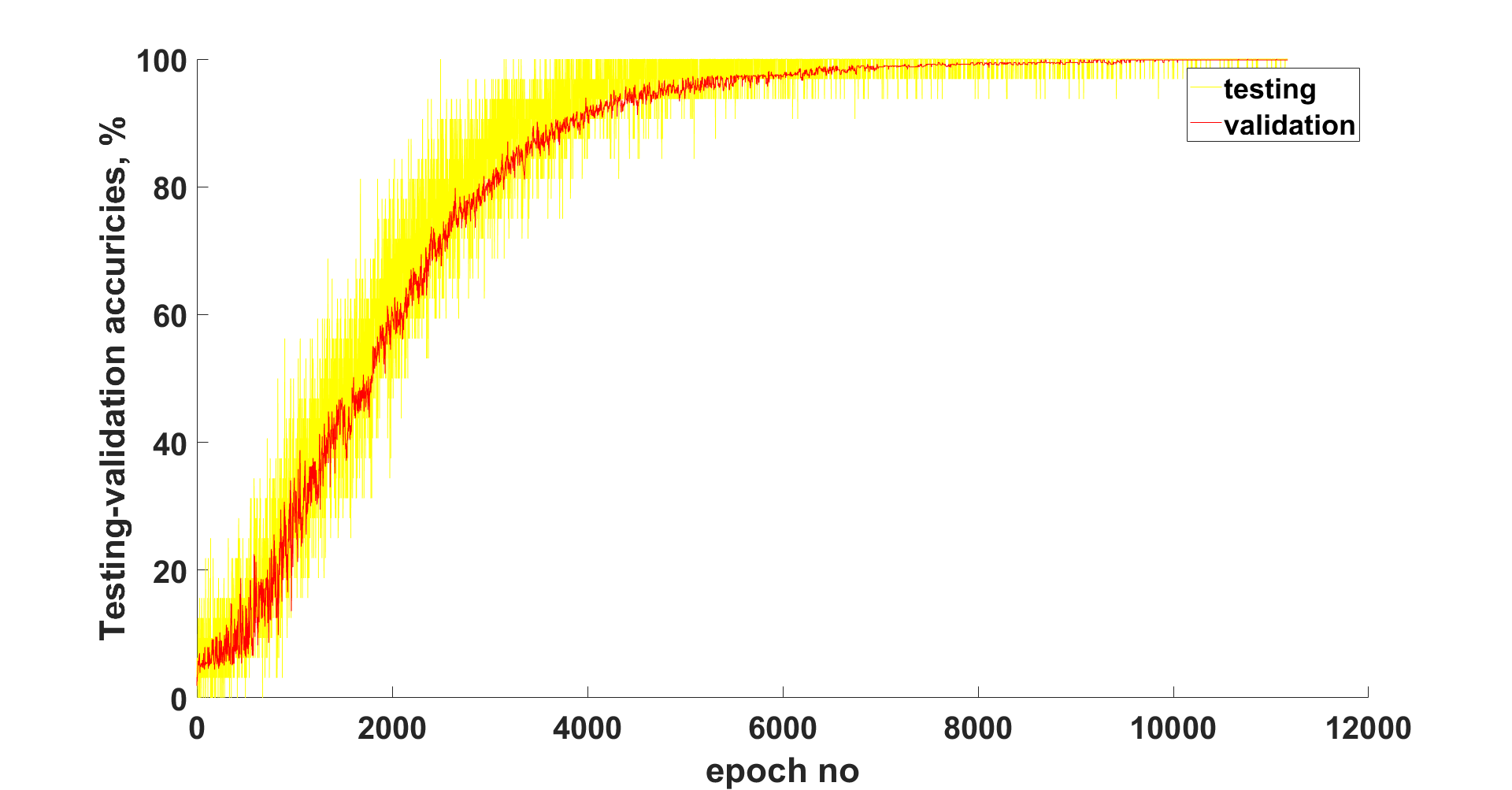}
	\caption{The performance of the Fig.~\ref{fig:ShallowNet} network. Learning profiles for the classification on categories `who drew', `who is drawn', and `age category'.}
	\label{fig:classification}
\end{figure}

\paragraph{Regression.}All the networks have predicted \textit{`age'} as a numeric value with testing and validation MAPE not exceeding 0.4\% (with MAE as a loss function). The performance of the `lightest' network is shown in Fig.~\ref{fig:regression}.
\paragraph{Classification.} Following factors and their combinations ---
\begin{itemize}
	\item `age',
	\item `age'+`who drew',
	\item `age'+`who drew'+`who is drawn',
	\item `age category',	
	\item `age in months'
\end{itemize}
--- were learned by all the presented network configurations (all having softmax as the last layer and cross-entropy  as a loss function)  with testing and validation accuracy (percentage of right guesses about categories) 100\% ! Look at Fig.~\ref{fig:classification}.
\paragraph{Freezeconnect.}Surprisingly good are not only the learning results. The calculations duration was, as a rule, 10 times and 30 times as small as the Keras network learning time extent for regression and classification, respectively.
Deepest IgNet structures with 6 and 7 layers required initially some more time. The duration was diminished by using newly introduced technique that we called \textit{freezeconnect}. Like known \textit{dropconnect}, it implies suspension of some randomly picked portion of weights from learning. But, contrary to it, freezeconnect \textit{does} include the frozen weights in relevant convolutions. Use of freezeconnect is illustrated by Fig.~\ref{fig:DeepestNet}. A practice like this has made the deepest networks almost as quick to teach as the shallower ones. This is likely because freezeconnect makes the network topology simpler and the loss function surface smoother. As a result, the convergence becomes more regular and fast.

\subsubsection{Why we do without neurons non-linear activation}
\label{sec:why}
One of the reasons for this was already pointed in subsection \ref{seq:esteem}~: better control of the loss function weights derivatives. Still, many enough experiments were made with various non-linear activators too. The best result here was performance of a network as in Fig.~\ref{fig:regression} but with \textit{bipolar sigmoid }activation of the second layer neurons. It learned much slower in both classification and regression cases, though was able to give 100\% classification accuracy. The regression MAPE, however, was about 4\%, i.~e., 10 times as high compared to the best results of IgNet. 

Periodically performed diagnostic evaluation of the weights derivatives (like in Fig.~\ref{fig:console}) showed pretty low derivatives absolute values --- this time\textit{ because the activators were approaching their saturation zone}. The problem that is really hard to deal with and that was earlier excluded due to using linear activation only. This experience is not necessarily `universal', though. It may happen that non-linear activators will become precious in future IgNet solutions.

\section{Conclusion}

Analysis of non-standard children\textquotesingle s drawings was thus presented. It was first done by means of widely known and generally used framework, then with newly developed CNN system which we called IgNet. IgNet\textquotesingle s solution was shown to take the edge of the results obtained with Keras--TensorFlow. Point once again IgNet\textquotesingle s  values that seem pivotal in having made such a result possible.
\begin{itemize}
	\item[\checkmark] IgNet can run in GPU or multithreaded CPU modes with adjustable number of threads, making the fastest of a concrete platform and performing great amount of calculations in a reasonable time period.
	\item[\checkmark] IgNet involves flexible construction of convolutions, possibly adjusting such parameters as zero padding, input padding, stride; biases may or may not learn together with weights.
	\item[\checkmark] Partial suspending is possible of neurons, weights, and biases from learning by \textit{dropout}, \textit{dropconnect}, as well as by newly introduced \textit{freezeconnect} for regularization or making network topology simpler if necessary.
    \item[$\diamondsuit$] IgNet includes a diagnostic tool evaluating values of weights partial derivative of the loss function. Controlling initial values of weights, one often can reach learning regime, far from both vanishing and explosion of gradient. This naturally leads to fast and precise learning making each neuron in the network as distinct as possible.
\end{itemize}	
This work will hopefully be a basis for getting more results --- in fields of unsupervised learning, binary/mixed data analysis, etc. In the meantime, the author is looking forward to any sort of feedback.
\\
\bibliographystyle{plain}
\bibliography{IgNet}

\begin{thebibliography}{1}

\bibitem{LeNet3}
R.~Bonnin.
\newblock {\em Building Machine Learning Projects with TensorFlow}.
\newblock Packt Publishing, 2016.

\bibitem{Dangeti}
P.~Dangeti.
\newblock {\em Statistics for Machine Learning}.
\newblock Packt Publishing, 2017.

\bibitem{DeepLearning}
I.~Goodfellow, Y.~Bengio, and {A. Courville}.
\newblock {\em Deep Learning}.
\newblock MIT Press, 2016.

\bibitem{LeNet1}
A.~Gulli.
\newblock {\em Deep Learning with Keras}.
\newblock Packt Publishing, 2017.

\bibitem{Heaton1}
J.~Heaton.
\newblock {\em Machine Learning: A Probabilistic Perspective}.
\newblock The MIT Press, Cambridge, 2012.

\bibitem{GPU}
D.~Kaeli, P.~Mistry, and {D. Schaa et al.}
\newblock {\em Heterogeneous Computing with OpenCL 2.0}.
\newblock Elsevier, 2015.

\bibitem{LeNet2}
Y.~S.~Resheff T.~Hope and {I. Lieder}.
\newblock {\em Learning TensorFlow: A Guide to Building Deep Learning Systems}.
\newblock O’Reilly, 2017.

\bibitem{CPU}
A.~Williams.
\newblock {\em C++ Concurrency in Action, Second Edition}.
\newblock Manning, 2019.

\end{thebibliography}
\end{document}